\documentclass[10pt,twocolumn,letterpaper]{article}

\usepackage[pagenumbers]{cvpr} 

\usepackage{graphicx}
\usepackage{amsmath}
\usepackage{amssymb}
\usepackage{booktabs}

\usepackage[pagebackref,breaklinks,colorlinks]{hyperref}

\usepackage[capitalize]{cleveref}
\crefname{section}{Sec.}{Secs.}
\Crefname{section}{Section}{Sections}
\Crefname{table}{Table}{Tables}
\crefname{table}{Tab.}{Tabs.}

\usepackage{multirow}
\usepackage{array} 

\usepackage{makecell}

\def\cvprPaperID{9} 
\def\confName{CVPR}
\def\confYear{2023}

\begin{document}

\title{Benchmarking the Physical-world Adversarial Robustness of Vehicle Detection}

\author{Tianyuan Zhang$^{1,2}$,Yisong Xiao$^{1,2}$,Xiaoya Zhang$^{1}$, Hao Li$^{1}$, Lu Wang$^{1}$\\
$^1$ School of Computer Science and Engineering, Beihang University, Beijing, China\\
$^2$ State Key Lab of Software Development Environment, Beihang University, Beijing, China\\
{\tt\small $\{$19373397, xiaoyisong, 20373300, 20373566, 20373361$\}$@buaa.edu.cn}}
\maketitle

\begin{abstract}
Adversarial attacks in the physical world can harm the robustness of detection models. Evaluating the robustness of detection models in the physical world can be challenging due to the time-consuming and labor-intensive nature of many experiments. Thus, virtual simulation experiments can provide a solution to this challenge. However, there is no unified detection benchmark based on virtual simulation environment. To address this challenge, we proposed an instant-level data generation pipeline based on the CARLA simulator. Using this pipeline, we generated the DCI dataset and conducted extensive experiments on three detection models and three physical adversarial attacks. The dataset covers 7 continuous and 1 discrete scenes, with over 40 angles, 20 distances, and 20,000 positions. The results indicate that Yolo v6 had strongest resistance, with only a 6.59\% average AP drop, and ASA was the most effective attack algorithm with a 14.51\% average AP reduction, twice that of other algorithms. Static scenes had higher recognition AP, and results under different weather conditions were similar. Adversarial attack algorithm improvement may be approaching its 'limitation'.
\end{abstract}

\section{Introduction}
\label{sec:intro}

\begin{figure}
    \centering
    \includegraphics[width=0.45\textwidth]{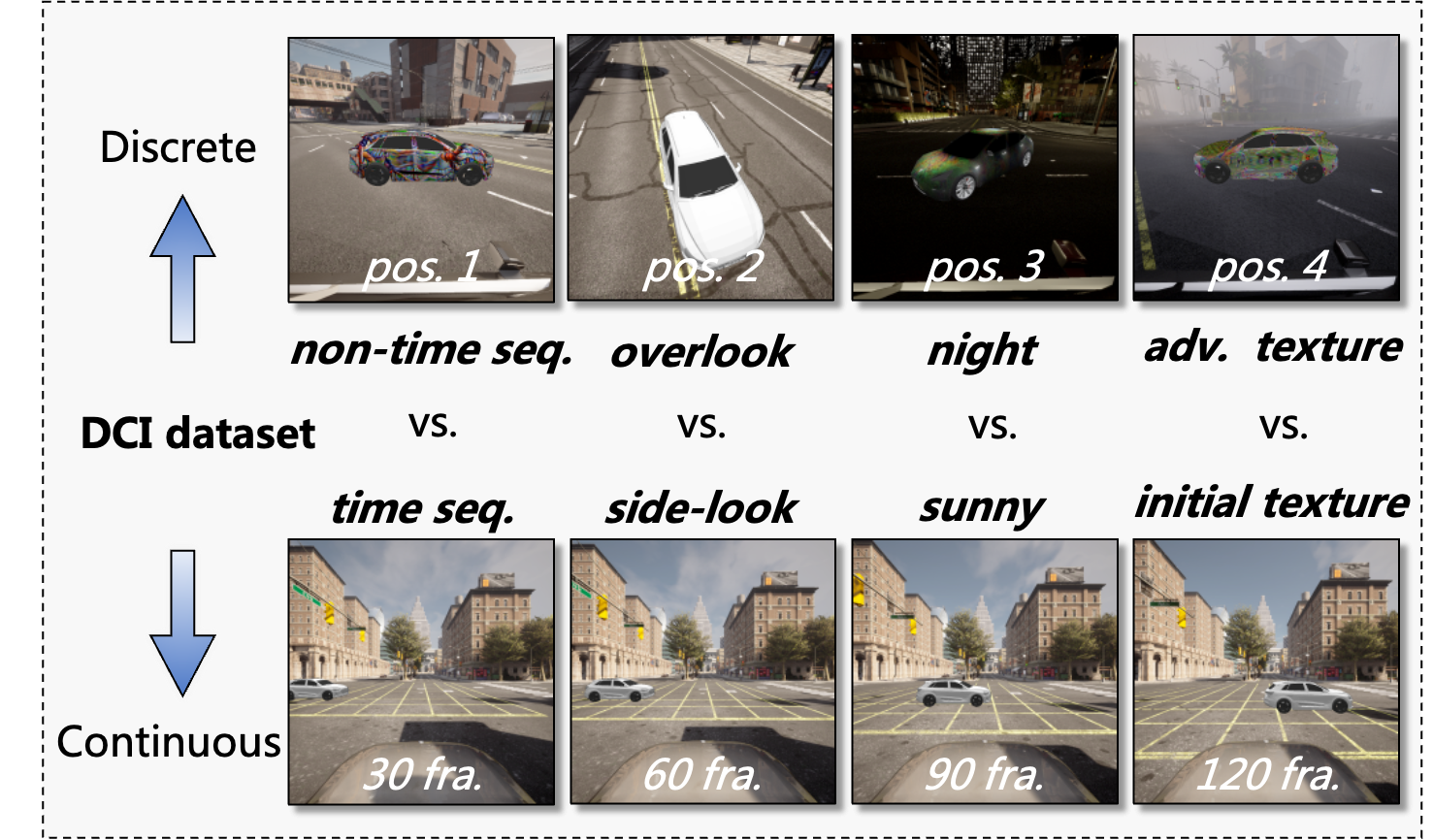}
    \caption{The Discrete and Continuous Instant-level Dataset (DCI): the discrete part aims to provide all-round coverage, while the continuous part is designed to test specific scenarios in greater depth.}
    \label{fig:DCIdataset}
\end{figure}

Detection models are vulnerable to adversarial perturbations, resulting in incorrect results. To overcome the time-consuming and labor-intensive problems of physical experiments, virtual simulation environments are gaining recognition as a valuable alternative, effectively addressing challenges such as inconvenient testing, difficult reproducibility, and high costs. Several adversarial attack algorithms  \cite{zhang2019camou,huang2020universal,wang2021dual, wang2022fca, zhang2022transferable} for vehicle detection scenarios have been proposed using CARLA simulators, revealing robustness issues. However, there is no widely accepted benchmark to support this research. To address this gap, we propose an instant-level scene generation pipeline based on CARLA and create the Discrete and Continuous Instant-level (DCI) dataset, covering various scenarios with different sequences, perspectives, weather, textures, and more. Figure \ref{fig:DCIdataset} illustrates different parts of the DCI dataset. Our main \textbf{contributions} are summarized as follows.

\begin{itemize}
\item[$\bullet$] We propose the DCI dataset to serve as a benchmark for evaluating the adversarial robustness of vehicle detection in the physical world.

\item[$\bullet$] We extensively evaluate three detection models and three adversarial attack algorithms using the DCI dataset, demonstrating the effectiveness of these attacks under various scenarios.
\end{itemize}

\section{Related Work}
\subsection{Adversarial Robustness Benchmark}
Several physical-world adversarial example generation methods have been proposed and demonstrated to be effective \cite{liu2019perceptual, liu2020spatiotemporal, liu2020bias, wang2021universal, wang2022defensive, liu2022harnessing,liu2023exploring}. However, they use different dataset for evaluation, which makes it difficult to conduct a comprehensive evaluation. To address this issue, several benchmarks have been proposed, including those by Dong \etal \cite{dong2020benchmarking} and Liu \etal \cite{liu2021training}. Tang \etal \cite{tang2021robustart} proposed the first unified Robustness Assessment Benchmark, RobustART, which provides a standardized evaluation framework for adversarial examples.

In the virtual simulation environment, several adversarial attack algorithms for vehicle recognition scenarios have been proposed\cite{zhang2019camou, wang2021dual, zhang2022transferable, wang2022fca} and shown to be effective. The CARLA simulator\cite{dosovitskiy2017carla} has been widely used in these studies due to its versatility and availability. However, the lack of a unified evaluation benchmark makes it difficult to compare and analyze the results. Establishing a benchmark is essential to promote the development of robust vehicle detection models.

\subsection{Virtual Environment of Vehicle Detection}
A series of vehicle detection-related simulators have been proposed. Simulators developed based on the Unity engine, such as LGSVL\cite{rong2020lgsvl}, and those developed based on the Unreal engine, such as Airsim\cite{shah2018airsim} and CARLA\cite{dosovitskiy2017carla}, all support camera simulation. Among them, the Airsim simulator focuses more on drone-related research, while compared with LGSVL, current research on adversarial security is more focused on the CARLA simulator\cite{wang2021dual,zhang2022transferable,wang2022fca}. CARLA is equipped with scenes and high-precision maps made by RoadRunner, and provides options for map editing. It also supports environment lighting and weather adjustments, as well as the simulation of pedestrian and vehicle behaviors.

Based on the above exploration, this study intends to use the CARLA autonomous driving simulator as the basic simulation environment to carry out research on the security analysis of autonomous driving intelligent perception algorithms.

\section{DCI Dataset: Instant-level Scene Generation and Design}

\subsection{Instant-level Scene Generation Pipeline}
For scenario generation, as mentioned earlier, we utilize the CARLA simulator as the underlying renderer and combine it with the Neural renderer\cite{kato2018neural} to balance feasibility and fidelity of the test. The CARLA renderer provides the highest fidelity but is non-differentiable, while the neural renderer ensures traceable gradients, facilitating further research.

\begin{figure}
    \centering
    \includegraphics[width=0.49\textwidth]{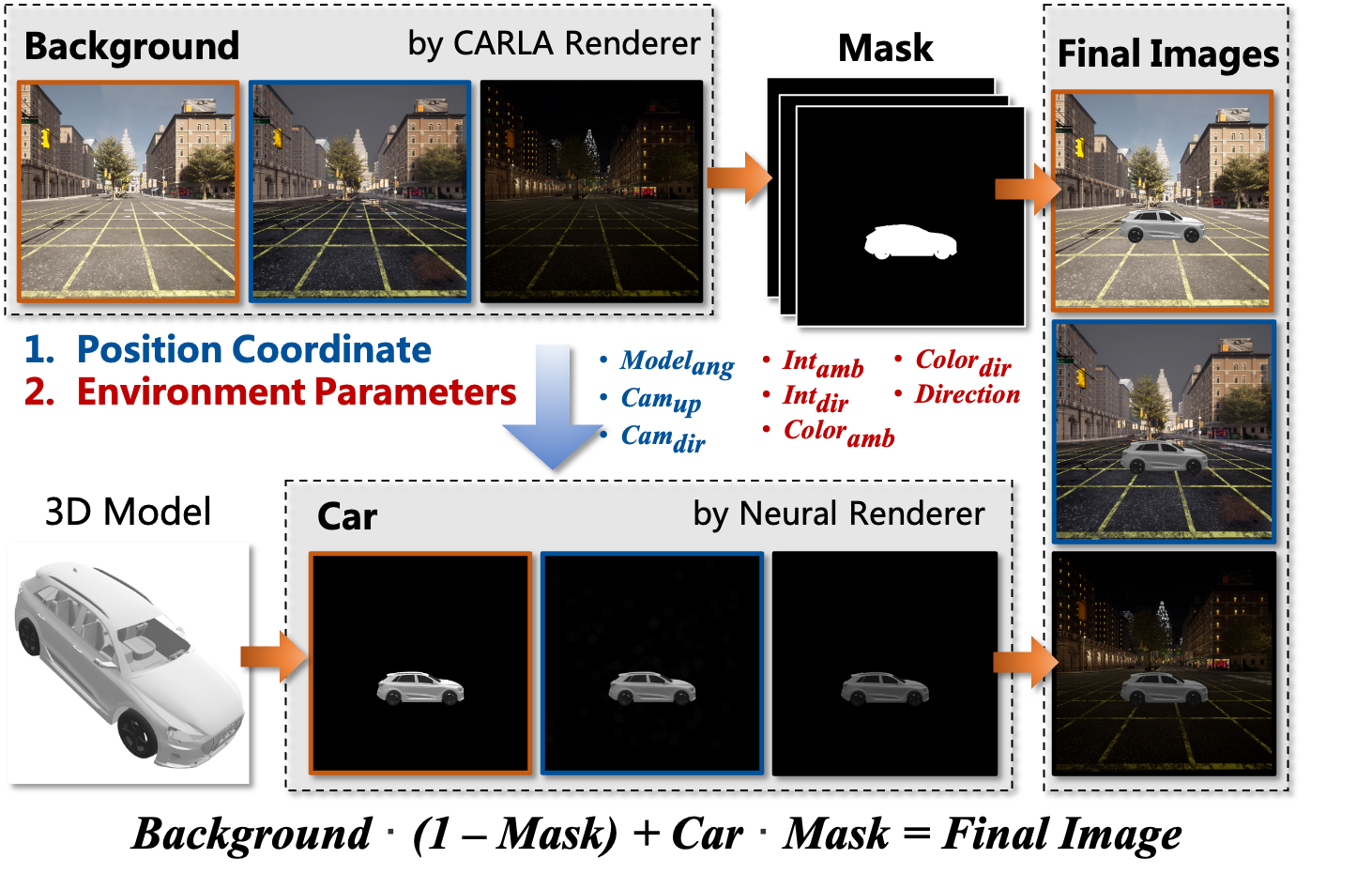}
    \caption{The pipeline and transferred parameters for instant-level scene generation}
    \label{fig:SceneGeneration}
\end{figure}

Previous methods only transferred position coordinates between the two renderers, resulting in a significant discrepancy between the synthesized images. To reduce the gap between the two renderers, we introduced transferred environmental parameters. The environmental parameters transferred between the two renderers are listed in Figure \ref{fig:SceneGeneration}.

Specifically, we use the CARLA simulator to first generate the \textbf{\textit{Background}} image and obtain the position coordinates $P_{co}$ and environment parameters $P_{en}$ using the simulator's built-in sensor. Next, we transfer $P_{co}$ and $P_{en}$ to the neural renderer. The Neural renderer then loads the 3D model and uses the received parameters to generate the \textbf{\textit{Car}} image. During the rendering process, we adjust the relevant settings of the neural renderer according to the sampling environment in CARLA to narrow the gap between the two renderers. We then use a \textbf{\textit{Mask}} to extract the background image and vehicle image respectively. After completing the pipeline, we obtain an instant-level scene. The framework of scene generation is shown in Figure \ref{fig:SceneGeneration}.

\begin{table*}
    \centering
    \caption{The AP of class Car on the DCI datasets}
    \resizebox{0.96 \textwidth}{!}{
    \begin{tabular}{*{14}{l}}
      \toprule
      \multirow{2}*{Scene} & \multirow{2}*{weather} & \multicolumn{3}{c}{initial} & \multicolumn{3}{c}{FCA} & \multicolumn{3}{c}{ASA}  &\multicolumn{3}{c}{DAS} \\
      \cmidrule(lr){3-5}\cmidrule(lr){6-8}\cmidrule(lr){9-11}\cmidrule(lr){12-14}
      & & $AP_{YOLOv3}$ & $AP_{YOLOv6}$ & $AP_{frcnn}$ &  $AP_{YOLOv3}$ & $AP_{YOLOv6}$ & $AP_{frcnn} $
      &  $AP_{YOLOv3}$ & $AP_{YOLOv6}$ & $AP_{frcnn}$ &  $AP_{YOLOv3}$ & $AP_{YOLOv6}$ & $AP_{frcnn}$ \\
      \midrule
      Overall & $Random$ & 65.37 & \textbf{73.39} & 56.81  &56.80 & 68.44  & 47.21  & \textbf{41.59} & 65.68  & 44.76 & 57.39 & 65.66 & 50.49 \\
      \hline
      \multirow{3}*{Traffic Circle} 
      & $ClearNoon$ & 77.04& \textbf{88.44} & 34.7 & 48.97 &64.43 & 23.9 &41.63& 76 & 25.93 &52.00& 51.35 & \textbf{7.85} \\
      & $ClearNight$ &39.02& \textbf{87.43} & 16.88 & 18.11 & 71.48 & 4.96 &24.43 & 79.8 & 9.06 &11.82 & 66.19 & \textbf{1.5} \\
      & $WetCloudySunset$ &74.03& 84.98 & 6.71 & 67.42 & 74.44 & \textbf{6.46} & 61.55 & 82.94 & 14.83 & 58.39 & 54.94 & 8.8 \\
      \hline
      \multirow{3}*{\textbf{Parking Lot}} 
      & $ClearNoon$ &16.86& 29.54 & \textbf{30.32} & 14.9 & 16.22 & 11.95 & \textbf{1.49} &13.81 & 10.74 & 13.14 & 15.66 & 16.06\\
      & $ClearNight$ &29.84& 37.99 & \textbf{44.98} & 20.54 & 26.75 & 16.78 & \textbf{8.54} & 15.81 & 11.68 & 20.79 & 23.07 & 25.88\\
      & $WetCloudySunset$ &22.39& \textbf{29.46} & 25.61 & 12.68 & 17.02 & 12.02 & 11.38 & 15.38 & \textbf{11.22} & 13.6 & 14.94 & 13.34\\
      \hline
      \multirow{3}*{Stationary A} 
      & $ClearNoon$ &66.47& 67.92 & \textbf{68.57} & 54.59 & 55.54 & 57.01 & \textbf{19.1} & 39.06 & 48.97 & 55.48 & 55.48 & 64.11\\
      & $ClearNight$ &66.02& \textbf{68.64} & 66.55 & 38.38&64.96 & 64.68 & \textbf{19.97} & 48.34 & 59.83 & 45.5 & 59.86 & 65.37\\
      & $WetCloudySunset$ &66.33& \textbf{70.4} & 68.56 & 48.61& 62.79 & 39.17 & \textbf{35.03} & 59.92 & 57.41 &45.94 & 59.51 & 62.76\\
      \hline
      \multirow{3}*{Straight Through A} 
      & $ClearNoon$ &89.52& 85.77 & \textbf{88.44} & 88.57 & 86.68 & 64.18 & \textbf{58.01} & 79.79 & 69.55 & 83.23 & 88.57 & 81.18\\
      & $ClearNight$ &79.12& 81.35 & \textbf{82.86} & 81.13 & 82.61 & 80.31 & \textbf{59.96} & 79.22 & 82.29 & 80.54 & 79.53 & 83.52\\
      & $WetCloudySunset$ &74.16& 80.36 & 74.29 &77.27 & 80.63 & 70.88 & \textbf{60.24} & \textbf{80.64} & 72.13 & 75.11 & 80.6 & 75.49\\
      \hline
      \multirow{3}*{Turning Left A} 
      & $ClearNoon$ &42.15& \textbf{67.42} & 16.31 & 58.7 & 82.28 & 14.12 & 55.76 & 73.89 & \textbf{6.52} & 69.58 & 78.41 & 14.38\\
      & $ClearNight$ &79.44& \textbf{79.46} & 22.75 & 77.31 & 80.39 & 18.34 & 71.1 & 77.91 & \textbf{15.16} & 79.06 & 79.23 & 19.52\\
      & $WetCloudySunset$ &22.47& 46.13 & \textbf{15.73} & 31.38 & \textbf{59.31} & 17.11 & 25.81 & 51.83 & 16.24 & 30.17 & 58.87 & 17.24\\
      \hline
      \multirow{3}*{Stationary B} 
      & $ClearNoon$ &99.59& \textbf{100} & \textbf{100} & 91.66 & 89.87 & 98.53 & \textbf{50.89} & 86.18 & 90.93 & 96.52 & 94.83 & 96.94\\
      & $ClearNight$ &98.77& \textbf{100} & \textbf{100} & 64.13 & \textbf{100} & 99.89 & \textbf{56.33} & 94.41 & 93.7 & 81.67 & 99.89 & 99.61 \\
      & $WetCloudySunset$ &98.07& \textbf{100} & \textbf{100} & 88.52 & 92.13 & 82.19 & \textbf{55.56} & 98.87 & 83.48 & 72.53 & 87.97 & 90.63\\
      \hline
      \multirow{3}*{Straight Through B}
      & $ClearNoon$ &75.08& 77.24 & \textbf{79.12} & 62.36 & 68 & 64.65 & \textbf{27.32} & 69.46 & 47.95 & 69.74 & 74.47 & 64.63\\
      & $ClearNight$ &79.77& 80.99 & 74.89 & 78.25 & 82.09 & 71.16 & 76.52 & 81.97 & \textbf{58.75} & 81.77 & \textbf{82.25} & 74.19\\
      & $WetCloudySunset$ &75.32& 77.82 & 75.71 & 69.27& \textbf{79.59} & 73.03 & \textbf{52.92} & 74.2 & 53.59 & 70.8 & 73.23 & 77.32\\
      \bottomrule
    \end{tabular}
    }
    
    \label{tab:results}
\end{table*}

\subsection{DCI Dataset Design}
The Discrete and Continuous Instant-level (DCI) dataset is designed to evaluate the performance of vehicle detection models in diverse scenarios. It can be divided into two parts that focus on different aspects.

The \textbf{Continuous} part of the DCI dataset comprises 7 typical scenes, each describing a real-life scenario that is widely used. To address the issue of irregular data distribution, we adopted a multi-perspective approach, including driver, UAV, and monitoring, to continuously sample real-world application backgrounds. To expand the coverage, we chose three different weather conditions to generate the dataset: \textit{ClearNoon}, \textit{ClearNight}, and \textit{WetCloudySunset}. This part of data set involves seven angles, distances and more than 2000 different positions.

The \textbf{Discrete} part of the DCI dataset is designed to expand coverage by selecting different maps, sampling distances, pitch angles, azimuth angles, and other parameters. We traverse the road positions in the map while fine-tuning environmental conditions such as lighting angle, lighting intensity, environmental haze, and particle density to meet the general test requirements. This part of data set involves 40 angles, 15 distances and more than 20000 different positions.


\section{Experiments and Evaluations}

\subsection{Experiment Settings}  
\textbf{Adversarial Attack Algorithm.} We employed four algorithms to generate adversarial examples: the initial texture, DAS algorithm \cite{wang2021dual}, FCA algorithm \cite{wang2022fca}, and ASA algorithm \cite{zhang2022transferable}. These algorithms were chosen based on their effectiveness in generating adversarial examples and their compatibility with the proposed method.

\textbf{Vehicle 3D Model.} We used the Audi E-Tron, a commonly used 3D model in previous studies, for our experiments. The model comprises 13,449 vertices, 10,283 vertex normals, 14,039 texture coordinates, and 23,145 triangles.

\textbf{Vehicle Detection Algorithm.} We evaluated the proposed method on three popular object detection algorithms: YOLO v3 \cite{redmon2018yolov3} YOLO v6\cite{li2022yolov6} and Faster R-CNN \cite{ren2015faster}. By selecting both single-stage and two-stage typical algorithms, we investigated the capability of the attack algorithm in the real world. The target class we chose is the car. We used the Average Precision (AP) as the evaluation metric to measure the performance of the detection algorithm on the test dataset.

\begin{figure*}
     \centering
     \begin{subfigure}[b]{0.33\textwidth}
         \centering
         \includegraphics[width=\textwidth]{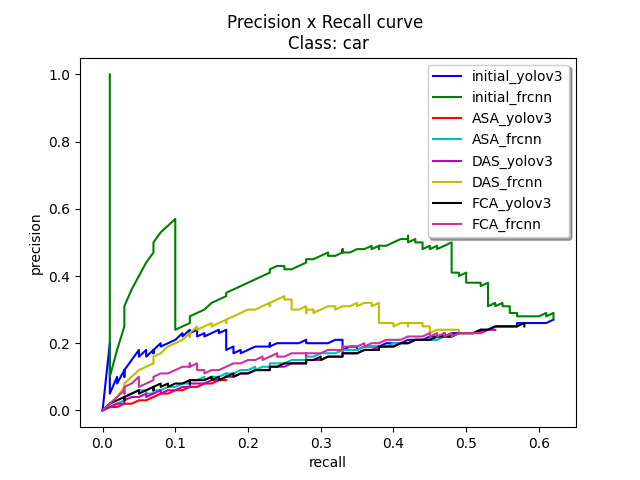}
         \caption{$ClearNoon$}
         \label{fig:ClearNoon}
     \end{subfigure}
     \begin{subfigure}[b]{0.33\textwidth}
         \centering
         \includegraphics[width=\textwidth]{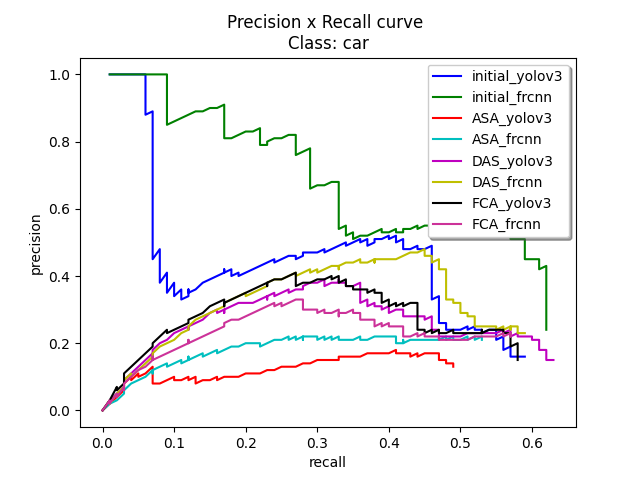}
         \caption{$ClearNight$}
         \label{fig:ClearNight}
     \end{subfigure}
     \begin{subfigure}[b]{0.33\textwidth}
         \centering
         \includegraphics[width=\textwidth]{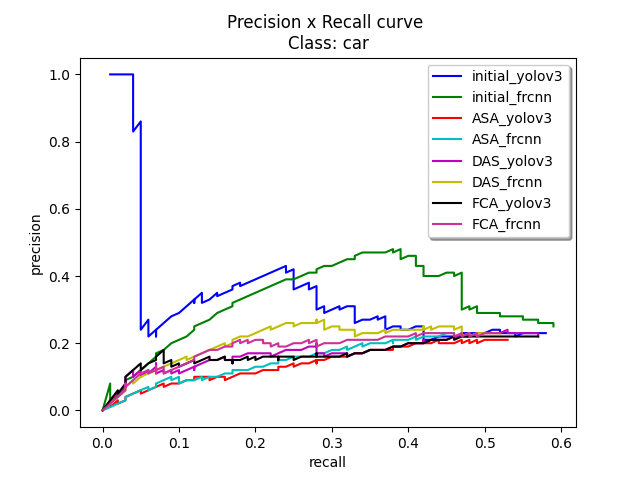}
         \caption{$WetCloudySunset$}
         \label{fig:WetCloudySunset}
     \end{subfigure} 
    \caption{The Precision-Recall chart illustrates the Park Lot scenario in three different weather conditions, demonstrating a similar distribution of values.}
    \label{fig:three graphs}
\end{figure*}

\subsection{Evalutaions}  
The results presented in Table \ref{tab:results} are analyzed as follows.

\textbf{Comparison of Model Robustness}. In almost all experiments, the YOLO v6 model showed the highest AP value, whereas YOLO v3 and Faster RCNN exhibited mixed results. However, after introducing adversarial perturbations, the average AP drop rates for the YOLO v3, YOLO v6 and Faster RCNN were 13.44\%, 6.79\%, and 9.32\%, respectively, with YOLO v6 exhibiting the strongest stability and YOLO v3 showing the weakest robustness.

\textbf{Comparison of Adversarial Attack Algorithms}. The corresponding AP drop values for FCA, ASA and DAS were 7.71\%, 14.51\%, and 7.34\%, respectively, with the ASA algorithm achieving almost twice the attack effect of the other algorithms. However, this attack effect strongly depends on the scenario. Although ASA performs best in most scenarios, both DAS and FCA outperform ASA in the Traffic Circle scenario.

\textbf{Comparison of Instance-level Scenes}. There is a significant difference in AP values between different scenes, with an average recognition AP of only 29.66\% in the Parking Lot scenario, while the AP in the Stationary B scenario reaches 99.6\%. After analyzing scenes with different AP values, it was found that scenes with low AP values correspond to situations where the observer is moving.

\textbf{Specific Scenarios Analysis.} We selected the Parking Lot scenario for further analysis, and the Precision-Recall curve is presented in Figure \ref{fig:three graphs}. The two highest lines correspond to the initial texture. For other adversarial textures, the data distribution shows similarities despite differences in values. Since the attack is not limited in scope, this suggests that different attacks may have certain limitations.

\section{Conclusion}
In the experiment, Yolo v6 showed the strongest resistance to attacks with an average AP drop of only 6.59\%. ASA was the most effective attack algorithm, reducing AP by an average of 14.51\%, twice that of other algorithms. Static scenes had higher recognition AP, and results in the same scene under different weather conditions were similar. Further improvement of adversarial attack algorithms may be reaching the 'limitation'.

{\small
\bibliographystyle{abbrv}
\bibliography{egbib}
}

\end{document}